\documentclass{llncs}

\usepackage{makeidx}
\usepackage{graphicx}
\usepackage{url}
\usepackage{algpseudocode}
\usepackage[ruled,section]{algorithm}
\usepackage{booktabs}
\usepackage{enumitem}
\usepackage{multirow}
\usepackage{amsmath}
\DeclareMathOperator{\diag}{diag}

\usepackage[usenames,dvipsnames]{xcolor}
\definecolor{darkgreen}{RGB}{0,128,0}

\begin{document}

\frontmatter

\pagestyle{headings}
\addtocmark{Hamiltonian Mechanics}

\title{Context-aware recommendations from implicit data via scalable tensor factorization\thanks{This paper is an extended version of our work published at ECML-PKDD'12\cite{itals_ecml}. Here we added two approximate variants of the main approach to address the scalability issues of ALS. We also extended the discussion with new experiments and comparisons, improved notation and clarified the description of algorithms, and restructured certain parts of the paper for the sake of better readability and completeness.}}

\author{Bal\'azs Hidasi\inst{1,}\inst{2,} \and Domonkos Tikk\inst{1}}

\authorrunning{B. Hidasi \& D. Tikk} 

\institute {Gravity R\&D Ltd. \and Budapest University of Technology and Economics
\email{\{balazs.hidasi,domonkos.tikk\}@gravityrd.com}}

\maketitle

\begin{abstract} \small\baselineskip=9pt
    Albeit the implicit feedback based recommendation problem---when only the user history is available but there are no ratings---is the most typical setting in real-world applications, it is much less researched than the explicit feedback case. State-of-the-art algorithms that are efficient on the explicit case cannot be automatically transformed to the implicit case if scalability should be maintained. There are few implicit feedback benchmark data sets, therefore new ideas are usually experimented on explicit benchmarks. In this paper, we propose a generic context-aware implicit feedback recommender algorithm, coined iTALS. iTALS applies a fast, ALS-based tensor factorization learning method that scales linearly with the number of non-zero elements in the tensor. We also present two approximate and faster variants of iTALS using coordinate descent and conjugate gradient methods at learning. The method also allows us to incorporate various contextual information into the model while maintaining its computational efficiency. We present two context-aware variants of iTALS incorporating seasonality and item purchase sequentiality into the model to distinguish user behavior at different time intervals, and product types with different repetitiveness.
Experiments run on six data sets shows that iTALS clearly outperforms context-unaware models and context aware baselines, while it is on par with factorization machines (beats 7 times out of 12 cases) both in terms of recall and MAP.

\end{abstract}

\section{Introduction}\label{sec:intr}

Recommender systems are information filtering algorithms that help users in information overload to find interesting items (products, content, etc). Users get personalized recommendations that contain typically a few items deemed to be of user's interest. The relevance of an item with respect to a user is predicted by recommender algorithms; items with the highest prediction scores are displayed to the user.

Recommender algorithms are usually sorted into two main approaches: the content based filtering (CBF) and the collaborative filtering (CF). CBF algorithms use user metadata (e.g. demographic data) and item metadata (e.g. author, genre, etc.) and predict user preference using these attributes, e.g. by matching the user profile containing the preferred item metadata of the user with item metadata \cite{LopsRSH11}. In contrast, CF methods do not use metadata, but only data of user--item interactions. Depending on the nature of the interactions, CF algorithms can be further classified into explicit and implicit feedback based methods. In the former case, users provide explicit information on their item preferences, typically in the form of user ratings. In the latter case, however, users express their item preferences only implicitly, as they regularly use an online system; typical implicit feedbacks are item views and purchases. CF algorithms proved to be more accurate than CBF methods, if sufficient preference data is available \cite{PilaRecsys09}.

CF algorithms can be classified into memory-based and model-based ones. The former are neighbor methods that make use of item or user rating vectors to define similarity, and they calculate recommendations as a weighted average of similar item or user rating vectors. In the last few years, model-based methods gained enhanced popularity, because they were found to be much more accurate in the Netflix Prize \cite{Netflix}, a community contest launched in late 2006 that provided the largest explicit benchmark data set (100M ratings) for a long time.

Model-based methods build generalized models that intend to capture user preference. The most successful approaches are the latent factor algorithms. These represent each user and item as a feature vector and the rating of user $u$ for item $i$ is predicted as the scalar product of their feature vectors. Matrix factorization (MF) methods approximate the partially known rating matrix, and they may differ in the model building, the learning method and the objective function. For learning MF methods may apply alternating least squares (ALS; \cite{BellkorICDM07}), gradient \cite{brismf} and coordinate descent method \cite{Recsys10}, conjugate gradient method \cite{Recsys11}, singular value decomposition \cite{KorenKDD08}, or a probabilistic framework \cite{Salak08}.

Explicit feedback based methods are able to provide accurate recommendations if enough ratings are available. In certain application areas, such as movie rental, travel applications, video streaming, users have motivation to provide ratings to get better service, better recommendations, or to award or punish a certain vendor. In general, however, users of an arbitrary online service do not tend to provide ratings on items even if such an option is available, because (1) when purchasing they have no information on their satisfaction (2) they are not motivated to return later to the system to rate. In such cases, user preferences can only be inferred by interpreting user actions (also called \emph{events}). For instance, a recommender system may consider the navigation to a particular product page as an implicit sign of preference for the item shown on that page \cite{RicciRSH11}. The user history specific to items are thus considered as implicit feedback on user taste. The interpretation of implicit feedback data may not necessarily reflect user satisfaction which makes the implicit feedback based preference modeling a difficult task. For instance, a purchased item could be disappointing for the user, so it might not mean a positive feedback. We can neither interpret missing navigational or purchase information as negative feedback, that is, such information is not available. Despite its practical importance, the implicit feedback problem has been less studied until recently. The proposed solutions typically modifies explicit feedback algorithms to be applicable for the implicit case.

The classical MF methods only consider user--item interaction (ratings or events) when building the model. However, we may have additional information related to items, users or events, which are together termed \emph{contextual information}, or briefly \emph{context}. Context can be, for instance, the time or location of recommendation, social networks of users, or user/item metadata \cite{AdomaviciusRecsys08}. Integrating context can help to improve recommender models. Tensor factorization have been suggested as a generalization of MF for considering contextual information for the explicit case \cite{KaratzogluRecsys10}. Factorization machines (FM) \cite{RendleSIGIR11} can be applied for the context-aware implicit case using subsampling for the negative feedbacks combined with a Bayesian objective function. Here we propose a tensor factorization method specifically developed for the context-aware implicit feedback problem.

The novelty of our work is fourfold: (1) we developed a fast tensor factorization method---coined iTALS---that can efficiently factorize huge tensors; (2) we adapted this general tensor factorization to the implicit recommendation task; (3) we introduce two approximate variants of iTALS using coordinate descent and conjugate gradient methods for learning; these enables faster training potentially at the cost of a higher loss function value in the optimization; (4) we present two specific implementations of iTALS that consider different contextual information. The first variant uses seasonality which was also used in \cite{KaratzogluRecsys10} for the explicit problem. The second algorithm applies sequentiality of user actions and is able to learn association rule like usage patterns. By using these patterns we can tell apart items or item categories having been purchased with different repetitiveness, which improves the accuracy of recommendations. To our best knowledge, iTALS is the first factorization algorithm that uses this type of information.

The paper is organized as follows. Section~\ref{sec:rel} briefly reviews related work in the field of recommendation systems on context-awareness, tensor factorization, and implicit feedback algorithms. In Section~\ref{sec:algo} we introduce our tensor factorization method and its application to the implicit recommendation task. Here we also introduce two approximate variants of iTALS. Section~\ref{sec:derived} shows two application examples of our factorization method: (1) we show how seasonality can be included in recommendations and (2) we discuss how a recommendation algorithm can learn repetitiveness patterns from the data set. Section~\ref{sec:numres} presents the results of our experiments, and Section~\ref{sec:conc} sums up our work and derives the conclusions. 
\subsection{Notation}\label{sec:not}
We will use the following notation in the rest of this paper:
\begin{itemize}[noitemsep,topsep=0pt,parsep=0pt,partopsep=0pt]
    \item $A\circ B\circ\ldots$: The Hadamard (elementwise) product of $A$, $B$, \ldots. The operands are of equal size, and the result's size is also the same. The element of the result at index $(i,j,k,\ldots)$ is the product of the element of $A$, $B$, \ldots\ at index $(i,j,k,\ldots)$. This operator has higher precedence than matrix multiplication in our discussion.
    \item $A_{i}$: The $i^{\rm th}$ column of matrix $A$.
    \item $A_{i_1,i_2,\ldots}$: The $(i_1,i_2,\ldots)$ element of tensor/matrix $A$.
    \item $K$: The number of features, the main parameter of the factorization.
    \item $D$: The number of dimensions of the tensor.
    \item $T$: A $D$ dimensional tensor that contains only zeroes and ones (preference tensor).
    \item $W$: A tensor with the same size as $T$ (weight tensor).
    \item $S_i$: The size of $T$ in the $i^{\rm th}$ dimension ($i=1,\ldots, D$).
    \item $N^+$: The number of ratings (explicit case); non-zero elements in tensor $T$ (implicit case).
    \item $M^{(i)}$: A $K\times S_i$ sized matrix. Its columns are the feature vectors for the entities in the $i^{\rm th}$ dimension.
    \item $A_{j_1,\ldots,j_{i-1},j,j_{i+1},\ldots,j_D}$ denotes an element of tensor $A$ where the index in the $i^{\rm th}$ dimension is fixed to $j$, and other indices are arbitrary.
\end{itemize} 
\section{Related work}\label{sec:rel}

Context-aware recommender systems \cite{AdomaviciusACMTIS05} emerged as an important research area in the last years and entire workshops are devoted to this topic on major conferences (CARS series started in 2009, \cite{CARS2009}; CAMRA in 2010, \cite{CAMRA2010}). The application fields of context-aware recommenders include among others: movie \cite{BogersCARS10}, music \cite{BaltrunasCARS09}, point-of-interest recommendation \cite{BaderCRR11}, and citation recommendation \cite{HeWWW10}. Context-aware recommender approaches can be classified into three main groups: pre-filtering, post-filtering and contextual modeling \cite{AdomaviciusRecsys08}. \cite{BaltrunasCARS09} proposed a pre-filtering approach by partitioned user profiles into \emph{micro-profiles} based on the time split of user event falls, and experimented with different time partitioning. Post-filtering ignores the contextual data at recommendation generation, but filters out irrelevant items (in a given context) or adjust recommendation score (according to the context) when the recommendation list is prepared; see a comparison in \cite{PannielloRecsys09}. The tensor factorization based solutions, including our proposed approach, falls into the contextual modeling category.

Tensor factorization (TF) incorporates contextual information into the recommendation model. Let us have a set of items, users and ratings (or events) and assume that additional context of the ratings is available (e.g. time of the rating). Having $C$ different contexts, the rating data can be cast into a $D=C+2$ dimensional tensor, $T$. The first dimension corresponds to users, the second to items and the subsequent $C$ dimensions $[3,\ldots,D]$ are devoted to contexts. TF methods approximate this tensor via lower rank approximation using RMSE based objective function. In \cite{KaratzogluRecsys10}, a sparse HOSVD \cite{hosvd} method is presented, which decomposes a $D$ dimensional sparse tensor into $D$ matrices and a $D$ dimensional tensor. If the size of the original tensor is $S_1\times S_2\times \cdots\times S_{D}$ and the number of features is $K$ then the size of the matrices are $S_1\times K$, $S_2\times K$, \dots, $S_{D}\times K$ and the size of the tensor is $K\times \cdots\times K$. The authors use gradient descent to learn the model. The complexity of one training iteration scales \emph{linearly} with the number of ratings ($N^+$) and \emph{cubically} with the number of features ($K$), which is a large improvement compared to the dense HOSVD's $O(K\cdot (S_1+\cdots+S_{D})^{D})$. A further improvement was proposed by \cite{RendleSIGIR11}: their factorization machine (FM) scales linearly \emph{both} $N^+$ and $K$. However, if the original tensor is large and dense like for the implicit recommendation task then neither method scales well, because $N^+=S_1\cdots S_D$.

For the implicit feedback problem, the naive minimization of the objective function is typically expensive; scales with the size of the user--item matrix. There are two dominant approaches to overcome this difficulty: (1) the trade-off solution that sacrifices the accuracy to some extent for computational efficiency by sampling the objective function; (2) the direct minimization of the objective function without sampling by decomposing the calculation to independent parts.

For direct minimization, the seminal work was proposed by \cite{HuICDM08}; their implicit ALS (iALS) applies an alternating least squares optimization and decomposes the derivatives of the objective function to user-dependent and user-independent parts, hence the complexity of a training iteration is reduced to scale \emph{linearly} with the number of positive feedbacks ($N^+$).

A faster, approximate version of iALS was proposed by \cite{Recsys10}, where the speedup was achieved by replacing the exact least squares solver by a coordinate descent method. The authors reported on a marginal loss of accuracy compared to the significant decrease in training time in their experiments. \cite{one_class_cf} proposed two prediction based frameworks for handling implicit feedback. The first one is similar to iALS, but it contains a naive ALS optimizer instead of a tuned one. The second one is based on negative example sampling. For ranking based objective function, \cite{commendo_track2} applied a stochastic gradient descent (SGD) optimizer on a sampled approximation of the objective function, while \cite{Recsys12} proposed a direct minimization for the same objective function with an appropriate decomposition of the derivatives. Another ranking based approach for implicit feedback is the Bayesian Personalized Ranking (BPR; \cite{bpr}), where objective function of BPR is derived from the Bayesian analysis of the problem. The optimization technique used for training is bootstrapping based SGD. For the implicit feedback problem (as suggested in \cite{Rendle12}), we combined the context-aware FM \cite{RendleSIGIR11} with BPR's bootstrapping learning. This method is used in our experiments for comparison. 
\section{ALS based fast tensor factorization}\label{sec:algo}

In this section we present iTALS, a general ALS-based tensor factorization algorithm that scales linearly with the non-zero element of a dense tensor (when appropriate weighting is used) and cubically with the number of features. This property makes our algorithm suitable to handle the context-aware implicit recommendation problem. At the end of this section we present approximate solutions for the ALS learning that have even better scaling properties in exchange for higher loss function values.

\begin{figure}[!h]
\centering
\includegraphics[width=2.4in]{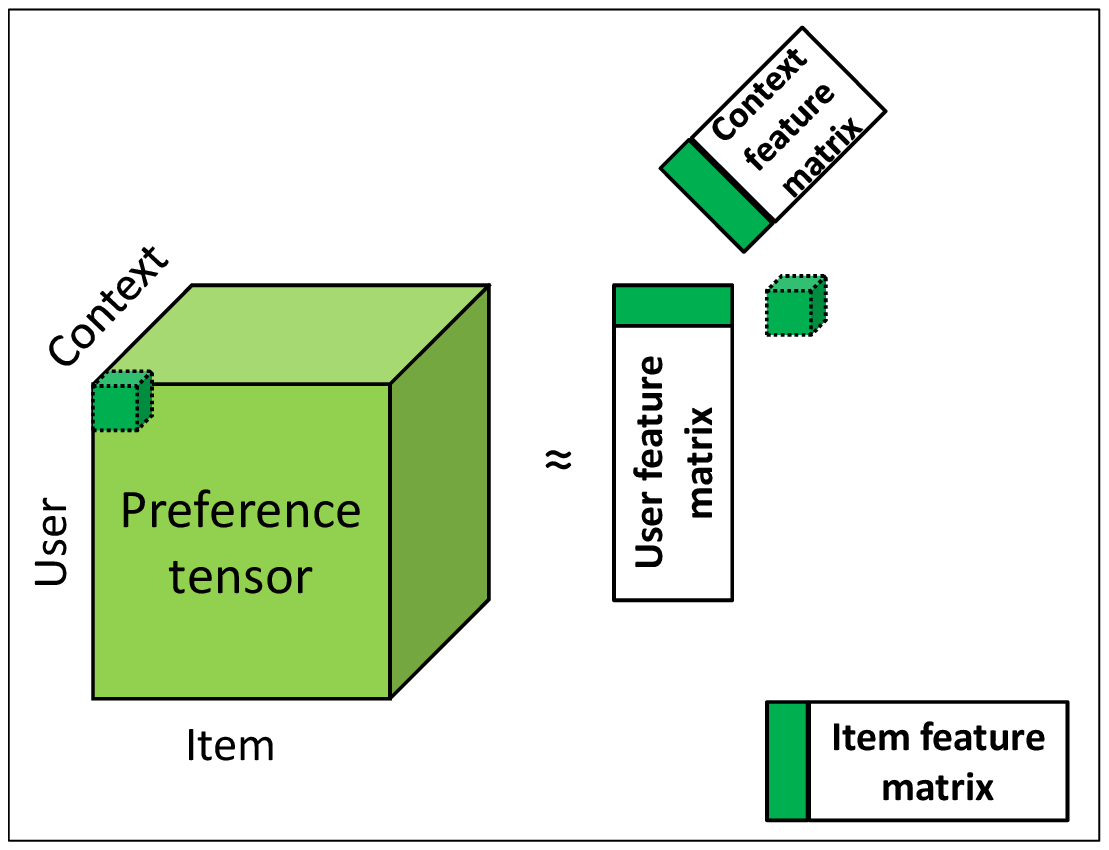}
\caption{Concept of the tensor decomposition in 3 dimension with the classical user--item--context setting.}
\label{fig:tensormodel}
\end{figure}

Let $T$ be a tensor of zeroes and ones and let $W$ contain weights to each element of $T$. Let $T_{u,i,c_1,\cdots,c_{D-2}}=1$ if user $u$ has (at least one) event on item $i$ while the context-state of $j^{\rm{th}}$ context dimension was $c_j$. Due to its construction, $T$ is very sparse. The weight matrix $W$ takes element $w_0$ if the corresponding element in $T$ is $0$, and is set to be greater than $w_0$ otherwise. A good choice of $W$ in practice---that we use in our experiments ---is $w_0=1$ and $w_t=100$. Instead of using the form of the common HOSVD decomposition ($D$ matrices and a $D$ dimensional tensor) our proposed model (see eq.~(\ref{eq:reconstruct})) approximates $T$ by a decomposition into $D$ matrices. The size of the matrices are $K\times S_1, K\times S_2, \ldots, K\times S_{D}$. (See figure~\ref{fig:tensormodel}.) The approximation of a given element of $T$ is the elementwise product of columns from $M^{(i)}$ low rank matrices:
\begin{equation}\label{eq:reconstruct}
\hat{T}_{i_1,i_2,\ldots,i_{D}}=1^TM^{(1)}_{i_1}\circ M^{(2)}_{i_2}\circ\cdots\circ M^{(D)}_{i_{D}}
\end{equation}
We want to minimize the loss function:
\begin{equation}\label{eq:loss}
L(M^{(1)},\ldots,M^{(D)})=\sum_{i_1=1,\ldots,i_{D}=1}^{S_1,\ldots,S_{D}}W_{i_1,\ldots,i_{D}}\left(T_{i_1,\ldots,i_{D}}-\hat{T}_{i_1,\ldots,i_{D}}\right)^2.
\end{equation}
The loss function $L$ is minimized by alternating least squares, that is, all but one of the $M^{(i)}$ matrices are fixed (without the loss of generality, next step are shown for $M^{(1)}$). Then $L$ is convex in the non-fixed variables. $L$ reaches its minimum in $M^{(1)}$, where its derivative with respect to $M^{(1)}$ is zero. Since the derivative of $L$ is linear in $M^{(1)}$, the columns of the matrix can be computed separately. That is for the $(i_1)^{\rm{th}}$ column of $M^{(1)}$:
\begin{equation}\label{eq:diff}
\begin{gathered}
0=\frac{\partial L}{\partial M^{(1)}_{i_1}} =
-2\underbrace{\sum_{i_2=1,\ldots,i_{D}=1}^{S_2,\ldots,S_{D}}W_{i_1,i_2,
\ldots,i_{D}}T_{i_1,\ldots,i_{D}}\left(M^{(2)}_{i_2}\circ\cdots
\circ M^{(D)}_{i_{D}}\right)}_{\mathcal{O}} +\\
2\underbrace{\sum_{i_2=1,\ldots,i_{D}=1}^{S_2,\ldots,S_{D}}W_{i_1,i_2,
\ldots,i_{D}}\left(M^{(2)}_{i_2}\circ\cdots\circ
M^{(D)}_{i_{D}}\right)\left(M^{(2)}_{i_2}\circ\cdots
\circ
M^{(D)}_{i_{D}}\right)^TM^{(1)}_{i_1}}_{\mathcal{I}}
\end{gathered}
\end{equation}
While $\mathcal{O}$ can be computed efficiently, the naive calculation of $\mathcal{I}$ is expensive (details are shown in Section~\ref{sec:complex}), therefore we transform $\mathcal{I}$ by using $W_{i_1,i_2,\ldots,i_{D}}=W'_{i_1,i_2,\ldots,i_{D}}+w_0$ and get:
\begin{equation}\label{eq:difftr}
\begin{gathered}
\mathcal{I}=\sum_{i_2=1,\ldots,i_{D}=1}^{S_2,\ldots,S_{D}}W'_{i_1,i_2,\ldots,i_{D}}\left(M^{(2)}_{i_2}\circ\cdots\circ M^{(D)}_{i_{D}}\right)\left(M^{(2)}_{i_2}\circ\cdots\circ M^{(D)}_{i_{D}}\right)^TM^{(1)}_{i_1}+\\
+\underbrace{w_0\sum_{i_2=1,\ldots,i_{D}=1}^{S_2,\ldots S_{D}}\left(M^{(2)}_{i_2}\circ\cdots\circ M^{(D)}_{i_{D}}\right)\left(M^{(2)}_{i_2}\circ\cdots\circ M^{(D)}_{i_{D}}\right)^T}_{\mathcal{J}}M^{(1)}_{i_1}
\end{gathered}
\end{equation}
The first sum in equation (\ref{eq:difftr}) is tractable because $W'_{i_1,i_2,\ldots,i_{D}}$ is zero for those indices where $T$ is zero. A similar decomposition step is applied in iALS \cite{HuICDM08}. $\mathcal{J}$ is the same for all columns of $M^{(1)}$ thus can be precomputed efficiently as follows.
\begin{equation}\label{eq:finaltr}
\begin{gathered}
\mathcal{J}=w_0\sum_{i_2=1,\ldots,i_{D}=1}^{S_2,\ldots,S_{D}}\left(M^{(2)}_{i_2}\circ\cdots\circ M^{(D)}_{i_{D}}\right)\left(M^{(2)}_{i_2}\circ\cdots\circ M^{(D)}_{i_{D}}\right)^T=\\
=w_0\underbrace{\left(\sum_{i_2=1}^{S_2}{M^{(2)}_{i_2}\left(M^{(2)}_{i_2}\right)^T}\right)}_{\mathcal{M}^{(2)}}\circ\cdots\circ\underbrace{\left(\sum_{i_D=1}^{S_D}{M^{(D)}_{i_D}\left(M^{(D)}_{i_D}\right)^T}\right)}_{\mathcal{M}^{(D)}}
\end{gathered}
\end{equation}
where we used that each element of $\mathcal{J}$ is computed as:
\begin{equation}\label{eq:righttr}
\begin{aligned}
\mathcal{J}_{j,k}=&w_0\left(\sum_{i_2=1,\ldots,i_{D}=1}^{S_2,\ldots,S_{D}}\left(M^{(2)}_{i_2}\circ\cdots\circ M^{(D)}_{i_{D}}\right)\left(M^{(2)}_{i_2}\circ\cdots\circ M^{(D)}_{i_{D}}\right)^T\right)_{j,k}=\\
=&w_0\sum_{i_2=1,\ldots,i_{D}=1}^{S_2,\ldots,S_{D}}\left(M^{(2)}_{j,i_2}\cdot\ldots\cdot M^{(D)}_{j,i_{D}}\right)\left(M^{(2)}_{k,i_2}\cdot\ldots\cdot M^{(D)}_{k,i_{D}}\right)=\\
=&w_0\left(\sum_{i_2=1}^{S_2}{M^{(2)}_{j,i_2}M^{(2)}_{k,i_2}}\right)\cdot\ldots\cdot\left(\sum_{i_D=1}^{S_D}{M^{(D)}_{j,i_D}M^{(D)}_{k,i_D}}\right)
\end{aligned}
\end{equation}

\begin{algorithm}[!h]
    \caption{Fast ALS-based tensor factorization for implicit feedback}\label{alg:iTALS}
    \textbf{Input:} {$T$: a $D$ dimensional $S_1 \times\cdots \times S_{D}$ sized tensor of zeroes and ones; $W$: a $D$ dimensional $S_1 \times\cdots \times S_{D}$ sized tensor containing the weights; $K$: number of features; $E$: number of epochs; $\lambda$: regularization coefficient} \newline
    \textbf{Output:} {$\{M^{(i)}\}_{i=1,\ldots, D}$} $K\times S_i$ sized low rank matrices \newline
    \textbf{procedure} \Call{iTALS}{$T$, $W$, $K$, $E$, $\lambda$}
	\begin{algorithmic}[1]
		\For{$i=1,\ldots,D$}
            \State $M^{(i)} \leftarrow $ Random $K\times S_i$ sized matrix
            \State $\mathcal{M}^{(i)} \leftarrow M^{(i)}(M^{(i)})^T$ \label{step:precomp}
        \EndFor
        \For{$e=1,\ldots,E$}
            \For{$i=1,\ldots,D$}\label{step:epochB}
                \State $C^{(i)} \leftarrow w_0\mathcal{M}^{(1)}\circ\cdots\circ \mathcal{M}^{(i-1)}\circ\mathcal{M}^{(i+1)}\cdots\circ \mathcal{M}^{(D)}$ \label{step:fast}
                \State $O^{(i)}\leftarrow 0$
                \For{$j=1,\ldots,S_i$}
                    \State $C^{(i,j)}\leftarrow C^{(i)}$
                    \State $O^{(i,j)}\leftarrow O^{(i)}$
                    \ForAll{$\{t \mid t=T_{j_1,\ldots,j_{i-1},j,j_{i+1},\ldots,j_D}, t\neq0\}$} \label{step:col_begin}
                        \State $W_t \leftarrow \diag(W_{j_1,\ldots,j_{i-1},j,j_{i+1},\ldots,j_D}-w_0)$
                        \State $v \leftarrow M^{(1)}_{j_1}\circ\cdots\circ M^{(i-1)}_{j_{1-1}}\circ M^{(i+1)}_{j_{i+1}}\circ\cdots\circ M_{j_D}^{(D)}$
                        \State $C^{(i,j)}\leftarrow C^{(i,j)}+ vW_tv^T$ \label{step:update1}
                        \State $O^{(i,j)}\leftarrow O^{(i,j)}+ W_tv$ \label{step:update2}
                    \EndFor \label{step:col_end}
                    \State $M^{(i)}_{j} \leftarrow (C^{(i,j)}+\lambda I)^{-1}O^{(i,j)}$ \label{step:reg}
                \EndFor
                \State $\mathcal{M}^{(i)} \leftarrow M^{(i)}(M^{(i)})^T$ \label{step:recomp}
            \EndFor\label{step:epochE}
        \EndFor
        \State \textbf{return} $\{M^{(i)}\}_{i=1,\ldots,D}$
\end{algorithmic}
\textbf{end procedure}
\end{algorithm}

The pseudocode of the proposed iTALS (Tensor factorization using ALS for implicit recommendation problem) is given in Algorithm~\ref{alg:iTALS}. The pseudocode follows the deduction above. In line \ref{step:precomp} we precompute $\mathcal{M}^{(i)}$ introduced in eq.~(\ref{eq:finaltr}). We create the column independent part of eq. (\ref{eq:difftr}) in line \ref{step:fast}.
We add the column dependent parts of eq. (\ref{eq:diff}) in lines \ref{step:col_begin}--\ref{step:col_end} and compute the desired column in line \ref{step:reg}. In this step we use regularization to avoid numerical instability and overfitting of the model. After each column of $M^{(i)}$ is computed, $\mathcal{M}^{(i)}$ is recomputed in line \ref{step:recomp}.

\subsection{Complexity}\label{sec:complex}
The computational cost of one epoch (lines \ref{step:epochB}--\ref{step:epochE}) is $O(DN^+K^2+K^3\sum_{i=1}^{D}{S_i})$. As shown above, $\mathcal{J}$ from eq.~(\ref{eq:finaltr}) is the same for each column, therefore its calculation is needed only once for each $M^{(i)}$ matrix, which takes $O(DK^2)$ time (see eq.~(\ref{eq:righttr})). To calculate $\mathcal{I}$ from eq.~(\ref{eq:difftr}) for the $j^{\rm{th}}$ column, we need the precomputed $\mathcal{J}$, as well as $O(N^+_jK^2)$ steps, where $N^+_j$ is the non-zero elements in $T$ with fixed $j$ dimension (the cardinality of the set in line~\ref{step:col_begin}). Although, the first sum of eq.~(\ref{eq:difftr}) runs over all entities of all but one dimension of $T$, $W'_{j_1,\ldots,j_{i-1},j,j_{i+1},\ldots,j_D}$ was constructed to be zero unless the respective element of $T$ is non-zero. To compute the partial derivative of the loss function with respect to a column of the non-fixed matrix $M^{(i)}$, we also need $\mathcal{O}$ (see eq.~(\ref{eq:diff})). It is calculated in $O(N^+_jK)$ time because most of $T_{j_1,\ldots,j_{i-1},j,j_{i+1},\ldots,j_D}$ are zero. Finally, we have to compute the actual feature vector that requires the inversion of a $K\times K$ matrix ($O(K^3)$). After all columns of the matrix are recomputed $\mathcal{M}^{(i)}$ also needs to be recomputed that takes $O(S_iK^2)$ time.

Summing these complexities for one $M^{(i)}$ matrix we need $O(S_iK^3+DK^2+\sum_{j=1}^{S_i}{N^+_j\left(K+K^2\right)}+S_iK^2)$ time that can be simplified to $O(N^+K^2+S_iK^3)$ (assuming that $D\ll S_i\ll N^+$, that is the case when the data is not too sparse). Summing this cost for all matrices we get $O(DN^+K^2+K^3\sum_{i=1}^{D}{S_i})$, which is cubical in $K$, the number of features, and linear in $N^+$, the number of non-zero elements of the tensor. In practical cases $DN^+ \gg \sum_{i=1}^{D}{S_i}$ (i.e. the data is not too sparse\footnote{$DN^+=\sum_{i=1}^{D}{S_i}$ means that we only have one event/example for each user, for each item and each context-state. In this case, CF method are not applicable due to sparseness.}), therefore the first term is dominant, thus the method scales \emph{quadratically} in $K$ for smaller $K$ values (also common in practice; see section~\ref{sec:runtimes} for actual running time measurements).

\subsection{Approximate solutions for ALS}\label{sec:appr}
Recall that except for the matrix inversion, our algorithm scales quadratically with $K$. The $DN^+K^2$ part dominates the $K^3\sum_{i=1}^{D}{S_i}$ part in the complexity when we use low-factor models, because usually $DN^+\gg\sum_{i=1}^{D}{S_i}$. The computation of the latter part can still requires a lot of time especially with higher $K$ values or more context dimensions. We introduce two approximate solutions and adapt them to iTALS to further reduce the time complexity of iTALS: iTALS-CD applies the coordinate descent learning for iTALS, while iTALS-CG adapts the conjugate gradient descent method. The adaptations are generalizations of the techniques proposed for matrix factorization in in \cite{Recsys10} and \cite{Recsys11}.

We will show that the approximate variants of iTALS can achieve lower running times (see Section~\ref{sec:runtimes}) in exchange for higher loss function values. Note that higher loss function values are not necessarily translated to lower accuracy in recommendations when one applies other, non-error based metrics (classification or ranking metrics) for the evaluation (see Section~\ref{sec:numres}).

\subsubsection{Coordinate descent}

The first approach for approximating the feature vector is to compute each of its coordinates separately, termed coordinate descent (CD). CD approximates the least squares solution of a $b=Ax$ linear system (seeking $x$). By fixing all but one feature and computing the remaining one, the matrix inversion can be avoided (it is reduced to a division) thus the computation time can be greatly reduced. Note that the while the solution provided by this method can be good enough, it does not converge to the least squares solution.

\begin{figure}[!h]
\centering
\includegraphics[width=4.9in]{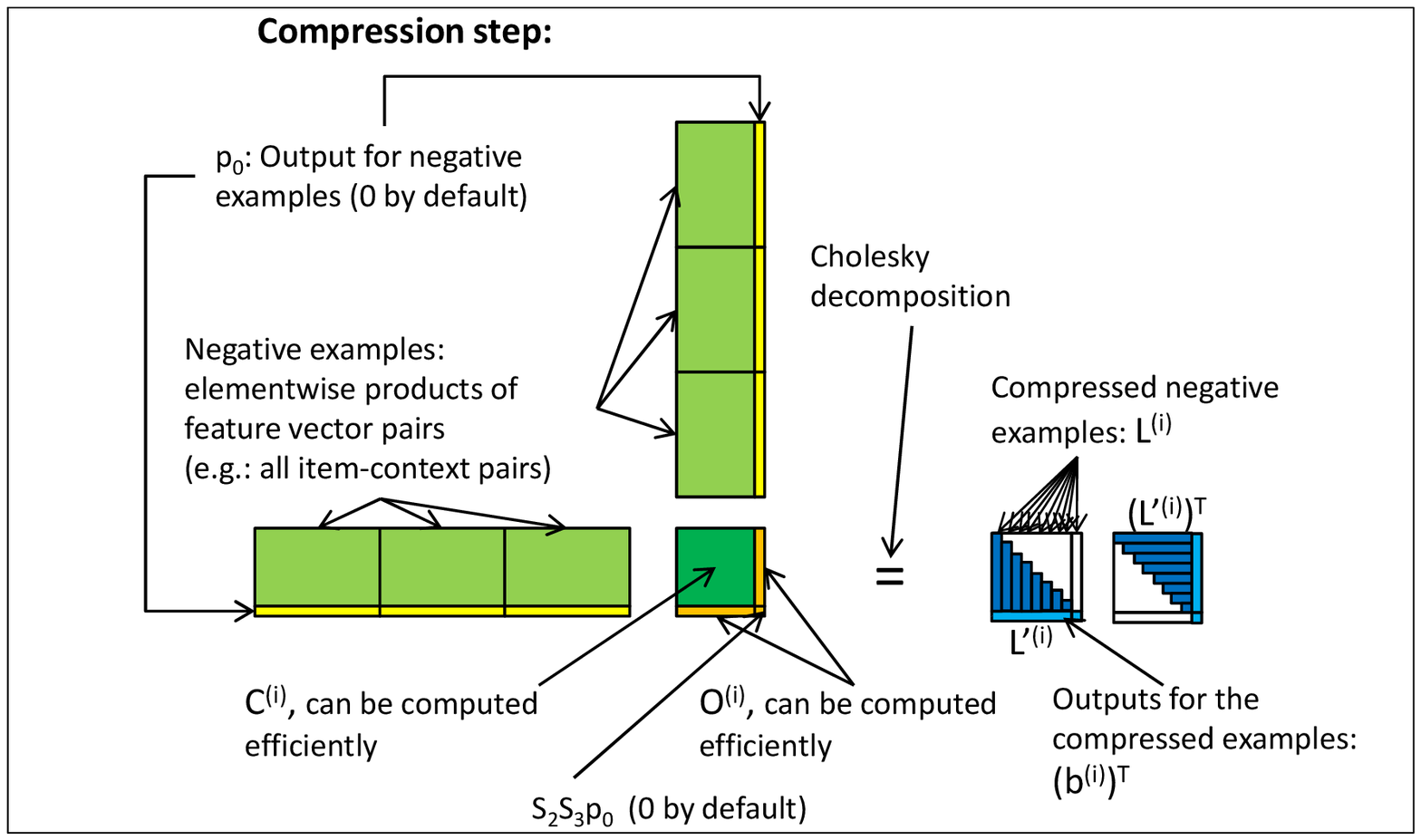}
\caption{Concept of the compression of negative examples in the 3 dimensional user--item--context setting.}
\label{fig:compressstep}
\end{figure}

The biggest difficulty to adapt CD to the iTALS framework is posed by the extremely high number of examples that corresponds  to the number of rows of $A$. This quantity is the product of the sizes of all but one dimension, that is $N_E=\prod_{j=1,j\neq i}^{D}{S_j}$, when the feature vector of the $i^{\rm{th}}$ dimension is sought. This quantity can be greatly reduced by compressing the information of the negative feedbacks, as shown next (also see figure~\ref{fig:compressstep}):

\begin{itemize}
    \item Compute the shared part of eq.~(\ref{eq:difftr}): $C^{(i)}$; and $O^{(i)}$. Recall that $C^{(i)}$ is in fact the covariance of all possible negative examples (i.e. the Hadamard product of all combinations of feature vectors from all but the $i^{\rm{th}}$ dimension). Similarly $O^{(i)}$ is the covariance with the output, but since missing events are represented by zeroes in the tensor, $O^{(i)}$ is a vector of zeroes.
    \item $O^{(i)}$ is appended to $C^{(i)}$ from the right and $\left(O^{(i)}\right)^T$ from the bottom. Thus we get a $(K+1)\times(K+1)$ sized matrix: $C'^{(i)}$. The $(K+1,K+1)$ element of $C'^{(i)}$ is set to $p_0\prod_{j=1,j\neg=i}^{D}{S_j}$, where $p_0$ is the value associated with the negative preference ($p_0=0$ by default). $C'^{(i)}$ is symmetric and positive definite. This step is needed because the input and the output must be compressed simultaneously.
    \item $C'^{(i)}$ is decomposed into $L'^{(i)}\left(L'^{(i)}\right)^T$ using Cholesky decomposition. $L'^{(i)}$ is a lower triangular matrix. The decomposition requires $O(K^3)$, but has to be computed only once per recomputing a feature matrix.
    \item The columns of $L'^{(i)}$ are the compressed negative examples. The first $K$ coordinates of a column form the example and the last coordinate is the output for that input.
\end{itemize}

The number of examples for the negative feedback was compressed into $K+1$ examples, that is shared for every feature vector of the $i^{\rm{th}}$ matrix. For the $j^{\rm{th}}$ feature vector we also need the positive feedback for the $j^{\rm{th}}$ entity, but their number is low ($N^+_j$), therefore the coordinate descent method can be computed efficiently.

\begin{algorithm}[!h]
    \caption{iTALS using coordinate descent for speedup}\label{alg:itals-cd}
    \textbf{Input:} {$T$: a $D$ dimensional $S_1 \times\cdots \times S_{D}$ sized tensor of zeroes and ones; $W$: a $D$ dimensional $S_1 \times\cdots \times S_{D}$ sized tensor containing the weights; $K$: number of features; $E$: number of epochs; $\lambda$: regularization coefficient; $N_I$: number of inner iterations of the CD method} \newline
    \textbf{Output:} {$\{M^{(i)}\}_{i=1,\ldots, D}$} $K\times S_i$ sized low rank matrices \newline
    \textbf{procedure} \Call{iTALS}{$T$, $W$, $K$, $E$, $\lambda$, $N_I$}
	\begin{algorithmic}[1]
		\For{$i=1,\ldots,D$}
            \State $M^{(i)} \leftarrow $ Random $K\times S_i$ sized matrix
            \State $\mathcal{M}^{(i)} \leftarrow M^{(i)}(M^{(i)})^T$
        \EndFor
        \For{$e=1,\ldots,E$}
            \For{$i=1,\ldots,D$}
                \State $C^{(i)} \leftarrow w_0\mathcal{M}^{(1)}\circ\cdots\circ \mathcal{M}^{(i-1)}\circ\mathcal{M}^{(i+1)}\cdots\circ \mathcal{M}^{(D)}$
                \State $O^{(i)}\leftarrow 0$\label{step:identical}
                \State $C'^{(i)}\leftarrow$ append $O^{(i)}$ to $C^{(i)}$ from right and $\left(O^{(i)}\right)^T$ from bottom \label{step:compress-begin}
                \State $C'^{(i)}_{K+1,K+1} \leftarrow 0$
                \State $L'^{(i)}\left(L'^{(i)}\right)^T \leftarrow$ \Call{Cholesky-decomposition}{$C'^{(i,j)}$}
                \State $L^{(i)}\leftarrow$ strip the last row of $L'^{(i)}$
                \State $b^{(i)}\leftarrow$ the last row of $L'^{(i)}$ transposed \label{step:compress-end}
                \For{$j=1,\ldots,S_i$}
                    \State $L^{(i,j)}\leftarrow L^{(i)}$
                    \State $b^{(i,j)}\leftarrow b^{(i)}$
                    \State $w^{(i,j)}\leftarrow $ vector of $w_0$ values; same length as $b^{(i,j)}$\label{step:wweight}
                    \ForAll{$\{t \mid t=T_{j_1,\ldots,j_{i-1},j,j_{i+1},\ldots,j_D}, t\neq0\}$} \label{step:add-example-begin}
                        \State $w_t' \leftarrow W_{j_1,\ldots,j_{i-1},j,j_{i+1},\ldots,j_D}-w_0$
                        \State $v \leftarrow M^{(1)}_{j_1}\circ\cdots\circ M^{(i-1)}_{j_{1-1}}\circ M^{(i+1)}_{j_{i+1}}\circ\cdots\circ M_{j_D}^{(D)}$
                        \State $L^{(i,j)}\leftarrow $ append $v$ to $L^{(i,j)}$ from right
                        \State $b^{(i,j)}\leftarrow $ append $1$ to $b^{(i,j)}$
                        \State $w^{(i,j)}\leftarrow $ append $w'_t$ to $w^{(i,j)}$
                    \EndFor \label{step:add-example-end}
                    \State $M^{(i)}_{j} \leftarrow$ \Call{Solve-weighted-CD}{$L^{(i,j)}$, $b^{(i,j)}$, $M^{(i)}_j$, $w^{(i,j)}$, $\lambda$, $N_I$} \label{step:solve-cd}
                \EndFor
                \State $\mathcal{M}^{(i)} \leftarrow M^{(i)}(M^{(i)})^T$
            \EndFor
        \EndFor
        \State \textbf{return} $\{M^{(i)}\}_{i=1,\ldots,D}$
\end{algorithmic}
\textbf{end procedure}
\end{algorithm}

Algorithm~\label{alg:itals-cd} shows the pseudocode for iTALS-CD. It is identical with algorithm~\ref{alg:iTALS} until line~\ref{step:identical}. In lines~\ref{step:compress-begin}--\ref{step:compress-end} the negative examples are compressed. We also need a weight vector because we optimize for weighted RMSE (line~\ref{step:wweight}). Updating steps of this matrix and vectors with positive examples are executed in lines~\ref{step:add-example-begin}--\ref{step:add-example-end}. The solution for $M^{(i)}_{j}$ is computed in line~\ref{step:solve-cd} using a weighted coordinate descent method (see Appendix). The signature of the solver is \Call{Solve-weighted-CD}{$A$, $b$, $x_0$, $w$, $\lambda$, $N_I$}, where the linear system is $A^Tx=b$, $x_0$ is an initial solution, the error is weighted by weight vector $w$ as ($\left|\left|w\circ\left(b-A^Tx\right)\right|\right|$), $\lambda$ is the regularization parameter and $N_I$ is the number of iterations.

The complexity of computing the $i^{\rm{th}}$ matrix consists of the following parts: computing $C^{(i)}$ in $O(K^2D)$, $L'^{(i)}$ in $O(K^3)$, the computation of all positive examples in $O(N^+K)$, optimizing using CD\footnote{The complexity of algorithm~\ref{alg:cd} is $O(N_EN_IK)$ that is $O\left((K^2+N^+_jK)N_I\right)$ in our case for one feature vector.} in $O(N_I(S_iK^2+N^+K))$ and finally recomputing $\mathcal{M}^{(i)}$ in $O(S_iK^2)$ time. This sums up to $O(K^3+N_IS_iK^2+N_IN^+K)$ for the $i^{\rm{th}}$ matrix (assuming that $D \ll N_IS_i$). Therefore the cost of one epoch is $O(DK^3+N_IK^2\sum_{i=1}^{D}{S_i}+DN^+N_IK)$. Comparing this to the complexity of iTALS ($O(DN^+K^2+K^3\sum_{i=1}^{D}{S_i})$) we can observe the followings. iTALS-CD also scales cubically in $K$, however, the coefficient is reduced from $\sum_{i=1}^{D}{S_i}$ to $D$. The other two terms are similar, however there is $N_IK^2$ and $N_IK$ in the place of $K^3$ and $K^2$. If $N_I \ll K$ and $D$ is low (that is the case in practice), for smaller $K$ values it scales \emph{linearly} in $K$ because $DN^+ \gg \sum_{i=1}^{D}{S_i}$.

\subsubsection{Conjugate gradient}

The conjugate gradient (CG; \cite{cg}) method is the state-of-the-art iterative method for solving $Ax=b$ type systems of linear equations, where $A$ is symmetric positive definite. The geometric interpretation of CG is that first a direction is selected in which the error can be reduced the most. In the following iterations the algorithm selects the best direction that is pairwise conjugate to every previous directions.

iTALS-CG approximates the feature vectors by replacing $M^{(i)}_{j}=(C^{(i,j)}+\lambda I)^{-1}O^{(i,j)}$ in line~\ref{step:reg} of algorithm~\ref{alg:iTALS} with $\Call{solveCG}{A, b, x_0, M}$ with $A=C^{(i,j)}+\lambda I$, $b=O^{(i,j)}$, $x_0=0$ and $M=\diag{(C^{(i,j)}+\lambda I)}$. The pseudocode of the conjugate gradient method is presented in the appendix (see algorithm~\ref{alg:cg}). The conjugate gradient method converges to the exact solution in at most $K$ steps. If $N_I=K$ it provides the exact solution, however it is often sufficient to run fewer inner iterations for a good solutions. The bottleneck of the CG method is the matrix-vector multiplication with $A$ and the inversion of $M$ in each iteration (see Appendix).

Here $A=C^{(i,j)}+\lambda I=C^{(i)}+\sum_{N^+_j}^{k=1}{v_kW_tv_k^T}+\lambda I$ and we use the Jacobi preconditioner ($M=\diag(A)=\diag{(C^{(i,j)}+\lambda I)}$).  Therefore it takes $O(N_IN^+_jK+N_IK^2)$ time to compute the feature vector when careful implementation is used. This sums up to $O(N_IN^+K+S_iN_IK^2)$ for recomputing one matrix instead of the $O(S_iK^3)$ complexity of the exact method. Therefore the total complexity of iTALS-CG is $O(DN^+N_IK+N_IK^2\sum_{i=1}^{D}{S_i})$. Note that for iTALS-CG $\mathcal{I}$ is not needed only $\mathcal{J}$ (line~\ref{step:update1} can be omitted from algorithm~\ref{alg:iTALS} when using the CG solver). Therefore the term $DN^+K^2$ can be omitted from the computation time as well. If $N_I \ll K$ iTALS-CG scales quadratically in the number of features (instead of cubically) in theory. In practice ($DN^+ \ll \sum_{i=1}^{D}{S_i}$) it scales \emph{linearly} (instead of quadratically) with the number of features for small $K$ values. However, if $N_I \approx K$ then its complexity is the same as of the exact iTALS. Since there are differences in the constant multipliers, iTALS-CG in fact can be slower than the exact iTALS in this case. 
\section{Context-aware iTALS algorithm}\label{sec:derived}
In this section we derive two specific algorithms from the generic iTALS method presented in Section~\ref{sec:algo}. The first method uses seasonality as context, the second considers the user history as sequential data, and learns meta-rules about sequentiality and repetitiveness.

\subsection{Seasonality}\label{sec:season}
Many application areas of recommender systems exhibit the seasonality effect, therefore seasonal data is an obvious choice for context \cite{LiuCAMRA10}. Strong periodicity can be observed in most of the human activities: as people have regular daily routines, they also follow similar patterns in TV watching at different time of a day, they do their summer/winter vacation around the same time in each year. Taking the TV watching example, it is probable that horror movies are typically watched at night and animation is more popular afternoon or weekend mornings. Seasonality can also be observed in grocery shopping or in hotel reservation data.

In order to consider seasonality, first we have to define the length of the season. Within a season we do not expect repetitions in the aggregated behavior of users, but we expect that at the same time offset in different seasons, the aggregated behavior of the users will be similar. The length of the season depends on the data. For example it is reasonable to set the season length to be 1 day for video-on-demand (VoD) consumption, however, this is not an appropriate choice for shopping data, where 1 week or 1 month is more justifiable. Having the length of the season determined, we need to create \emph{time bands} (bins) in the seasons. These time bands are the possible context-states. Time bands specify the time resolution of a season, which is also data dependent. We can create time bands with equal or different length. For example, every day of a week are time bands of equal length, but 'morning', 'around noon', 'afternoon', 'evening', 'late evening', 'night' could be of different length. Obviously, these two steps require some a-priori knowledge about the data or the recommendation problem, but iTALS is not too sensitive to minor deviations related to the length and the resolution of the season.

In the next step, events are assigned to time bands according to their time stamp. Thus, we can create the (user, item, time band) tensor. We factorize this tensor using the iTALS algorithm and we get feature vectors for each user, for each item and for each time band. When a recommendation is requested for user $u$ at time $t$, first the time band of $t$ is determined and then the preference value for each item using the feature vector of user $u$ and the feature vector of time band $tb_t$ is calculated.

\subsection{Sequentiality}\label{sec:meta}
Recommendation algorithms often recommend items from categories the user likes. For example if the user often watches horror movies then the algorithm will recommend her horror movies. This phenomenon is even stronger if time decay is applied giving higher weights to recent events. Pushing newer events can increase accuracy, because similar items will be recommended. Such a consideration can be beneficial in some application fields, like VoD recommendation, but will fail in cases where repetitiveness in user behavior with respect to items can not be observed. A typical example for that is related to household appliance products: if a user buys a TV set and then she gets further TV sets recommended, she will not probably purchase another one. Instead, complementary or related goods are rather suitable to recommend, for example, DVD players or external TV-tuners. On the other hand, the purchase of a DVD movie does not exclude at all the purchase of another one. Whether recommendation of similar items is reasonable, depends on the nature of the item and the behavior of the user. Next, we propose an approach to integrate the repetitiveness of purchase patterns into the latent factor model.

Using association rules is a possible approach to specify item purchase patterns. Association rules \cite{Agrawal93} are often used to determine which products are bought frequently together and it was reported that in certain cases association rule based recommendations yield the best performance \cite{JamesRecsys10}. In our setting, we can extract purchase patterns from the data using association rule mining on the subsequent user events within a given time window.
There are two possibilities: we can generate category--category rules, or category--item rule, thus having usage patterns:

\begin{enumerate}
\item if a user bought an item from category $A$ then she will buy an item from category $B$ next time, or
\item if a user bought an item from category $A$ then she will buy an item $X$ next time.
\end{enumerate}

We face, however, with the following problems, when attempting to use such patterns in recommendations: (1) the parameter selection (minimum support, minimum confidence and minimum lift) influences largely the performance, their optimization may be slow; (2) rules with negated consequents (e.g. bought from $A$ will not buy from $B$) are not found at all; (3) with category--category rules one should devise further weighting/filtering to promote/demote the items in the pushed category; (4) the category--item rules are too specific therefore either one gets too many rules or the rules will overfit.

We show how repetitiveness related usage patterns can be efficiently integrated into recommendation model using the iTALS algorithm. Let us now consider the \emph{category of last purchased item} as the context for the next recommendation. The tensor has again three dimensions: users, items and item categories. The $(i,u,c)$ element of the tensor means that user $u$ bought item $i$ and the user's latest purchase (before buying $i$) was an item from category $c$. Using the examples above: the user bought a given DVD player after the purchase of a TV set. After factorizing this tensor we get feature vectors for the item categories as well. These vectors act as weights in the feature space that reweight the user--item relations. For example, assuming that the first item feature means ``having large screen'' then the first feature of the TV category would be low as such items are demoted. If the second item feature means ``item can play discs'' then the second feature of the TV category would be high as these items are promoted.

The advantage of this method is that it learns the usage patterns from the data globally by producing feature vectors that reweight the user--item relations. One gets simple but general usage patterns using the proposed solution that integrates seamlessly into the common factorization framework: no post-processing is required to define promotional/demotional weights/filters.

We can generalize the concept described above to take into account several recent purchases. We could create a $D=C+2$ dimensional tensor, where the $[3,\ldots,D]$ dimensions would represent the item categories of the last $C$ purchases, but the resulting tensor would be very sparse as we increase $C$. Instead we remain at a three dimensional tensor but we set simultaneously $C$ item categories to $1$ for each user--item pair. We may also decrease the weights in $W$ for those additional $C-1$ cells as they belong to older purchases. Thus we may control the effect of previous purchases based on their recency. When recommending, we have to compute the (weighted) average of the feature vectors of the corresponding categories and use that vector as the context feature vector. 
\section{Experiments}\label{sec:numres}
We used six data sets to evaluate our algorithm. Five of them contain genuine implicit feedback data (LastFM 1K, \cite{lastfm1k}; TV1, TV2, \cite{tv1_tv2}, and 2 proprietary), while the last one is an implicit variant of the MovieLens 10M explicit feedback data \cite{Movielens}. The properties of the data sets are summarized in Table~\ref{tab:data}. The column ``Multi'' shows the average multiplicity of user--item pairs in the training events.\footnote{This value is 1.0 at three data sets: for MovieLens, it is due to transformation of ratings into implicit feedback, while TV1 and TV2 data might have been filtered for duplicate events.} The train--test splits are time-based: the first event in the test set is after the last event of the training set. The length of the test period was selected to be at least one day, and depends on the domain and the frequency of events. We used the artists as items in LastFM. MovieLens ratings of 4.5 or above were transformed into positive implicit feedback, lower ratings were discarded. We remark that the results for MovieLens are mainly included for the sake of reproducibility; the focus of our study is genuine implicit feedback data.

\begin{table}[!h]
\centering
\caption{Main properties of the data sets}\label{tab:data}
\medskip
{\small
\begin{tabular}{@{}l@{\hskip2mm}l@{\hskip4mm}r@{\hskip2mm}r@{\hskip2mm}r@{\hskip2mm}r@{\hskip4mm}r@{\hskip2mm}l@{}}
\toprule
\multirow{2}{*}{\textbf{Dataset}} & \multirow{2}{*}{\textbf{Domain}}& \multicolumn{4}{c}{\textbf{Training set}}& \multicolumn{2}{c}{\textbf{Test set}} \\
&& \textbf{\#Users}& \textbf{\#Items}& \textbf{\#Events}& \textbf{Multi}& \textbf{\#Events} & \textbf{Length}\\
\midrule
Grocery & E-grocery & 24947 & 16883 & 6238269 & 3.0279 & 56449 & 1 month \\
TV1 & IPTV & 70771 & 773 & 544947 & 1.0000 & 12296 & 1 week \\
TV2 & IPTV & 449684 & 3398 & 2528215 & 1.0000 & 21866 & 1 day \\
VoD & IPTV/VoD & 480016 & 46745 & 22515406 & 1.2135 & 1084297 & 1 day \\
LastFM & Music & 992 & 174091 & 18908597 & 21.2715 & 17941 & 1 day \\
MovieLens & Movies & 69878 & 10681 & 2112790 & 1.0000 & 3112 & 1 week \\
\bottomrule
\end{tabular}}
\label{tab:tab}
\end{table}

Our primary evaluation metric is recall@20. Recall is defined as the ratio of relevant recommended items and relevant items. An item is considered relevant for a user if there is an event in the test data with the given user and item. Recall does not take into account the position of an item on the recommendation list. We estimate that users are exposed to 20 recommendations in average during a visit (e.g. 4 pageviews, 5 items per recommendation), therefore we choose cutoff at 20. Mean Average Precision (MAP) was used as a secondary evaluation metric. Unlike recall, MAP considers the order of the recommended items as well, thus prefers methods that put the relevant items at the beginning of the recommendation list. We also used a cutoff value of 20 for MAP (denoted as MAP@20). While MAP@20 measurements contain valuable information on the performance of the recommender algorithms, we find recall more appealing. In a practice recommended items are usually randomly selected from the first $N$ elements of the ranked item list. $N$ is small but larger than the number of recommendation boxes (e.g.: $N=20$). Our interest is that the user clicks one of the items, but it is irrelevant whether it was the first or the $N^{\rm{th}}$ item in our ranking. Therefore recall@N suits the offline evaluation of recommender algorithms from the practical viewpoint more than MAP@N.

\subsection{Performance of iTALS}

We determined the seasonality for each data set, that is, the periodicity of patterns observed in the data. For \emph{Grocery} we defined a week as the season and the days of the week as the time bands. The argument here is that people usually do shopping on weekly or biweekly basis and that shopping habits differ on weekends and weekdays. One day was used as season for the other five data sets with 4 hour intervals as the time bands, because households watch different types of programs at different time of the day. We note that one can optimize the length of time bands but this is beyond the scope of the current paper.

In the next experiment we used item sequentiality as context; the context state of an event is the item of the previous event of the same user.

We compared the two iTALS variants (seasonal and sequential data) to the basic iALS as well as to a context-aware baseline for implicit feedback data. This method, referred as \textit{implicit CA (iCA) baseline}, is the composite of several iALS models. For each context state we train a model using only the events with the appropriate context, e.g., with Grocery we train 7 models for the 7 time bands. The context of the recommendation request (e.g. day of week) selects the model for the prediction. This baseline treats context-states independently. Due to its long running time we used iCA only with seasonality, as \#(time bands) $\ll$ \#(preceding items).

iTALS is also compared to the implicit version of the FM method \cite{RendleSIGIR11}. In contrast to the 3-way model of iTALS ($D$-way in general), FM uses a full pairwise model, i.e. user--item + user--context + item--context. The training of the implicit FM is executed using BPR criterion with bootstrapped stochastic gradient. Since this option is not available in the original FM library \cite{Rendle12}, we use our own implementation in this comparison.

\begin{table}[!h]
\centering
\caption{Recall@20 and MAP@20 values for all data sets with iALS, iCA and iTALS using factorization with 20 features. The differences are significant at $p<0.01$ (paired t-test). Best results are typeset in bold.}
\medskip
{\tiny
\begin{tabular}{lcrrrrrr}
\toprule
\multirow{2}{*}{\textbf{Dataset}} & \multirow{2}{*}{\textbf{iALS}} & \multicolumn{2}{c}{\textbf{iCA baseline}} & \multicolumn{2}{c}{\textbf{iTALS}} & \multicolumn{2}{c}{\textbf{iTALS}} \\
&& \multicolumn{2}{c}{\textbf{(seasonality)}}& \multicolumn{2}{c}{\textbf{(seasonality)}}& \multicolumn{2}{c}{\textbf{(sequentiality)}}\\
\midrule
\multicolumn{8}{c}{\textbf{Recall@20}}\\
\midrule
Grocery & 0.0647 & 0.0783 & (20.98\%) & 0.1062 & (63.96\%) & \textbf{0.1203} & \textbf{(85.86\%)}\\
TV1 & 0.1195 & 0.1173 & (-1.84\%) & 0.1371 & (14.77\%) & \textbf{0.1415} & \textbf{(18.45\%)}\\
TV2 & \textbf{0.2160} & 0.1861 & (-13.84\%) & 0.1794 & (-16.96\%) & 0.2041 & (-5.52\%)\\
VoD & 0.0633 & 0.0807 & (27.42\%) & 0.0994 & (56.94\%) & \textbf{0.1035} & \textbf{(63.47\%)}\\
LastFM & 0.0229 & 0.0272 & (19.22\%) & 0.0339 & (48.29\%) & \textbf{0.0872} & \textbf{(281.46\%)}\\
MovieLens & 0.0932 & 0.0999 & (7.24\%) & \textbf{0.1019} & \textbf{(9.31\%)} & 0.1009 & (8.28\%)\\
\midrule
\multicolumn{8}{c}{\textbf{MAP@20}}\\
\midrule
Grocery & 0.0820 & 0.1220 & (48.73\%) & 0.1618 & (97.25\%) & \textbf{0.1695} & \textbf{(106.72\%)}\\
TV1 & 0.0306 & 0.0352 & (15.09\%) & 0.0430 & (40.52\%) & 0.0418 & (36.68\%)\\
TV2 & 0.0539 & 0.0463 & (-14.13\%) & 0.0488 & (-9.57\%) & 0.0670 & (24.30\%)\\
VoD & 0.0326 & 0.0710 & (118.03\%) & 0.0814 & (149.96\%) & 0.0602 & (84.79\%)\\
LastFM & 0.0236 & 0.0367 & (55.81\%) & 0.0420 & (78.01\%) & \textbf{0.1575} & \textbf{(567.88\%)}\\
MovieLens & 0.0340 & 0.0497 & (46.33\%) & \textbf{0.0623} & \textbf{(83.38\%)} & 0.0433 & (27.47\%)\\
\bottomrule
\end{tabular}}
\label{tab:results}
\end{table}

\begin{table}[!h]
\centering
\caption{Recall@20 and MAP@20 values for iTALS and FM using factorization with 20 features. The differences are significant at $p<0.01$ (paired t-test).}
\medskip
{\tiny
\begin{tabular}{lrrrrrr}
\toprule
\multirow{2}{*}{\textbf{Dataset}} & \textbf{FM} & \textbf{iTALS} & \textbf{(Diff)} & \textbf{FM} & \textbf{iTALS} & \textbf{(Diff)} \\
& \multicolumn{3}{c}{\textbf{(seasonality)}} & \multicolumn{3}{c}{\textbf{(sequentiality)}} \\
\midrule
\multicolumn{7}{c}{\textbf{Recall@20}}\\
\midrule
Grocery & 0.1032 & 0.1062 & (2.88\%) & 0.1039 & 0.1203 & (15.86\%)\\
TV1 & 0.1572 & 0.1371 & (-12.78\%) & 0.1235 & 0.1415 & (14.62\%)\\
TV2 & 0.2217 & 0.1794 & (-19.06\%) & 0.2649 & 0.2041 & (-22.96\%)\\
VoD & 0.0369 & 0.0994 & (169.20\%) & 0.1152 & 0.1035 & (-10.14\%)\\
LastFM & 0.0656 & 0.0339 & (-48.34\%) & 0.0605 & 0.0872 & (44.15\%)\\
MovieLens & 0.0996 & 0.1019 & (2.26\%) & 0.0980 & 0.1009 & (2.95\%)\\
\midrule
\multicolumn{7}{c}{\textbf{MAP@20}}\\
\midrule
Grocery & 0.1475 & 0.1618 & (9.68\%) & 0.1393 & 0.1695 & (21.73\%)\\
TV1 & 0.0469 & 0.0430 & (-8.33\%) & 0.0336 & 0.0418 & (24.44\%)\\
TV2 & 0.0681 & 0.0488 & (-28.43\%) & 0.0971 & 0.0670 & (-30.96\%)\\
VoD & 0.0227 & 0.0814 & (257.89\%) & 0.0914 & 0.0602 & (-34.20\%)\\
LastFM & 0.0908 & 0.0420 & (-53.78\%) & 0.0991 & 0.1575 & (58.87\%)\\
MovieLens & 0.0374 & 0.0623 & (66.85\%) & 0.0421 & 0.0433 & (2.98\%)\\
\bottomrule
\end{tabular}}
\label{tab:vsfm}
\end{table}

Every algorithm has three common parameters: the number of epochs, the regularization parameter and the number of features. The number of epochs was set to 10 as the head of ranked recommendation lists hardly change after 10 epochs. The regularization was proportional to the support of the given item/user/context and the ratio was optimized on a validation data set prior to training the model on the whole training data. We did not use any other heuristics (like time decay) to focus on the pure performance of the algorithms. We ran experiments with 20 and 40 features because of the practical importance of low factor models; also typical in the literature \cite{KorenKDD08,PilaRecsys09}. Since results were similar we showcase the 20-feature experiment.

Table~\ref{tab:results} shows the recall@20 and MAP@20 values for all data sets. The percentage values in parentheses show the improvement over the context unaware iALS algorithm. With the exception of the TV2 data set the incorporation of context information substantially improves the results. Even the baseline method increases recall@20 by 20\% on average. The iTALS algorithm improves the results of the baseline by 15\%--35\% that is a total of 15\%--65\% improvement over iALS. The increase in MAP@20 is even higher. This means that iTALS recommends more relevant items and that those items are ranked higher on the recommendation lists.

When sequentiality is used as context the improvement mainly depends on the average multiplicity of user--item pairs, because such data sets provide more examples to specific cases of sequentiality. For LastFM, a user--item pair is present $\approx21$ times on average and the recall is improved by more than 280\%. But it is important to note that even on data sets where there are absolutely no repeating items (e.g. TV1), sequentiality can increase performance. This is due that sequences of two items can still be observed multiple times in different user histories. However, the increase is smaller, because the behavior of users can be described less accurately by sequentiality. In certain cases, we have no information on whether the same item was consumed multiple times.

The only data set on which context aware methods achieved worse results than iALS is the \emph{\emph{TV2}} data set. The problem is not specific to iTALS, the context aware baseline also worsens the recall. This phenomenon highlights the importance of the proper context selection. Context aware algorithms are efficient for two reasons: (1) different sets of items can be recommended to the same user under different conditions, thus the more specific needs of the users can be met; (2) a good context information separates items and/or users well and so makes it easier for the algorithm to learn the user--item relations more accurately. As for TV2, the standardized season and sequentiality definition did not yield better results than the baseline, however, with specific context-states learnt from the given data set, the performance of iTALS can be improved. A thorough investigation of this direction is beyond the scope of our paper.

iTALS proved to be consistently more accurate than FM in 7 out of 12 experiments both in terms of recall and MAP (see Table~\ref{tab:vsfm}). One can observe large differences in both metrics in favor of either method depending on the data set. The characterization of problems when iTALS or FM consistently outperforms the other is beyond the scope of this paper, thus left for future research.

\subsection{Relation of content and time band lengths}
The effect of different context dimensions was examined in the next experiment. The LastFM and VoD databases were used to train the iTALS model with different seasonality information. The length of the season was kept fixed to one day. However the experiments were conducted using time bands of different length (8, 6, 4, 2, 1 hours, 30 and 10 minutes). The time bands are placed evenly in the 24 hours of the season, meaning that their length is equal, the first one stars at 0:00 and the last one ends at 24:00. This also means that the number of time bands is different for the experiments. (Regularization was optimized separately for each experiment using a validation set.)

\begin{figure*}[!h]
\centering
\includegraphics[width=0.9\hsize]{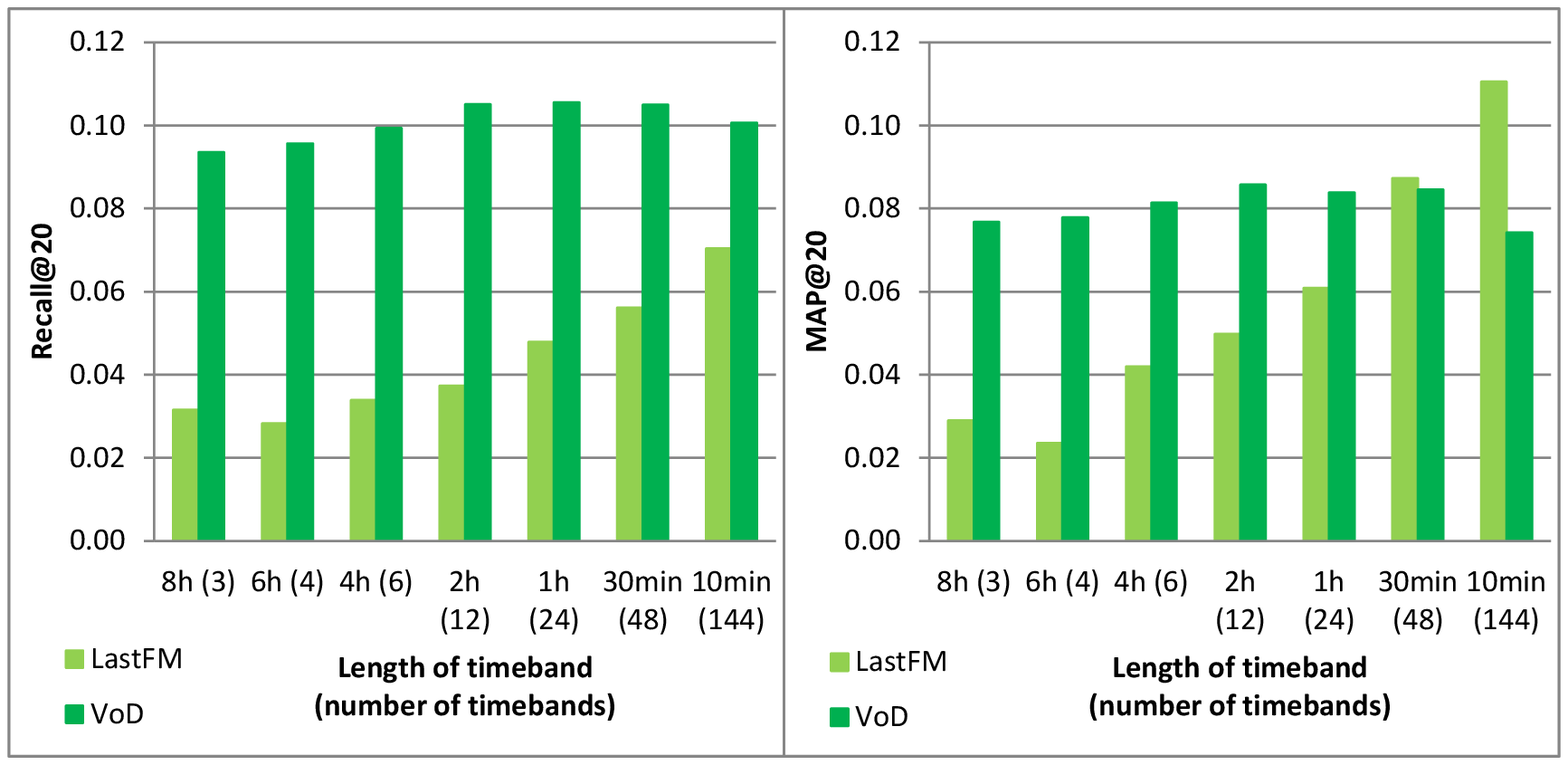}
\caption{Recall@20 and MAP@20 values of the iTALS trained on the LastFM and VoD datasets using different seasonality information.}
\label{fig:seasoneffect}
\end{figure*}

Figure~\ref{fig:seasoneffect} depicts the results of the experiments. The effect of the time band length is smaller on VoD than on LastFM. For VoD the best choices are 0.5--2 hour time bands with respect to recall, and an increase of $\approx6\%$ over the default 4 hour length (Table~\ref{tab:results}) can be observed. Remark that longer time bands (6--8 hours) cannot differentiate well enough the user behaviors at different time. MAP@20 is the best for the 2 hours long time bands. For LastFM, the shorter time band length, the better recall@20 and MAP@20 values are obtained. Here the best results are for 10 minutes time band length.

The length of content has a significant effect on the optimal choice of the time band length.
For VoD, the content length ranges between 20 and 120 minutes and the most common length are 30--45 minutes (episodes), while the music content (LastFM), length of the songs are much shorter, typically few minutes. Results suggest that the optimal time band length should be around 1.5--3 times of the typical item consumption time, with equally placed time bands.

\subsection{Comparison of training algorithms}\label{sec:runtimes}
We introduced three methods to learn the model in Section~\ref{sec:algo}. In addition to the original ALS learning scheme, we proposed two approximate variants: the coordinate descent (CD) and the conjugate gradient (CG). Table~\ref{tab:cdcg} shows the results of the three algorithms on all data sets with both seasonality and sequentiality. The results for all three methods are often quite similar. The approximate methods sometimes can slightly beat the original ALS, because the learning algorithms optimize for weighted RMSE but recall is the primary evaluation metric. Although, we can observe correlation between lower wRMSE and higher recall, it is not one-to-one relation.

\begin{table}[!h]
\centering
\caption{Recall@20 and MAP@20 values for ALS, CD and CG learning; Number of inner iterations is 2 for both CD and CG.}
\medskip
{\footnotesize
\begin{tabular}{lrrrrrrr}
\toprule
\multirow{2}{*}{\textbf{Dataset}} & \multicolumn{3}{c}{\textbf{iTALS season}}& \multicolumn{3}{c}{\textbf{iTALS seq.}} \\
& \textbf{ALS} & \textbf{CG} & \textbf{CD} & \textbf{ALS} & \textbf{CG} & \textbf{CD} \\
\midrule
\multicolumn{7}{c}{\textbf{Recall@20}}\\
\midrule
Grocery & 0.1062 & 0.1078 & 0.1053 & 0.1203 & 0.1192 & 0.1189\\
TV1 & 0.1371 & 0.1322 & 0.1361 & 0.1415 & 0.1405 & 0.1347\\
TV2 & 0.1794 & 0.1758 & 0.1811 & 0.2041 & 0.2044 & 0.2050\\
LastFM & 0.0339 & 0.0383 & 0.0381 & 0.0872 & 0.0862 & 0.0872\\
VoD & 0.0994 & 0.0996 & 0.0994 & 0.1035 & 0.1018 & 0.1017\\
MovieLens & 0.1019 & 0.0987 & 0.0999 & 0.1009 & 0.1009 & 0.1009\\
\midrule
\multicolumn{7}{c}{\textbf{MAP@20}}\\
\midrule
Grocery & 0.1618 & 0.1690 & 0.1623 & 0.1695 & 0.1683 & 0.1692\\
TV1 & 0.0430 & 0.0429 & 0.0423 & 0.0418 & 0.0416 & 0.0413\\
TV2 & 0.0488 & 0.0476 & 0.0483 & 0.0670 & 0.0637 & 0.0668\\
LastFM & 0.0420 & 0.0476 & 0.0726 & 0.1575 & 0.1743 & 0.2486\\
VoD & 0.0814 & 0.0816 & 0.0813 & 0.0602 & 0.0605 & 0.0593\\
MovieLens & 0.0623 & 0.0639 & 0.0673 & 0.0433 & 0.0433 & 0.0433\\
\bottomrule
\end{tabular}}
\label{tab:cdcg}
\end{table}

CD and CG have an additional hyperparameter over ALS, the number of inner iterations. We expect that the accuracy improves with more iteration at the cost of a slower learning. The main motivation behind using approximate algorithms is to speed up the training process. We found $N_I=2$ to be a fair trade-off between speed and performance for both methods.

\begin{figure}[!h]
\centering
\includegraphics[width=0.8\hsize]{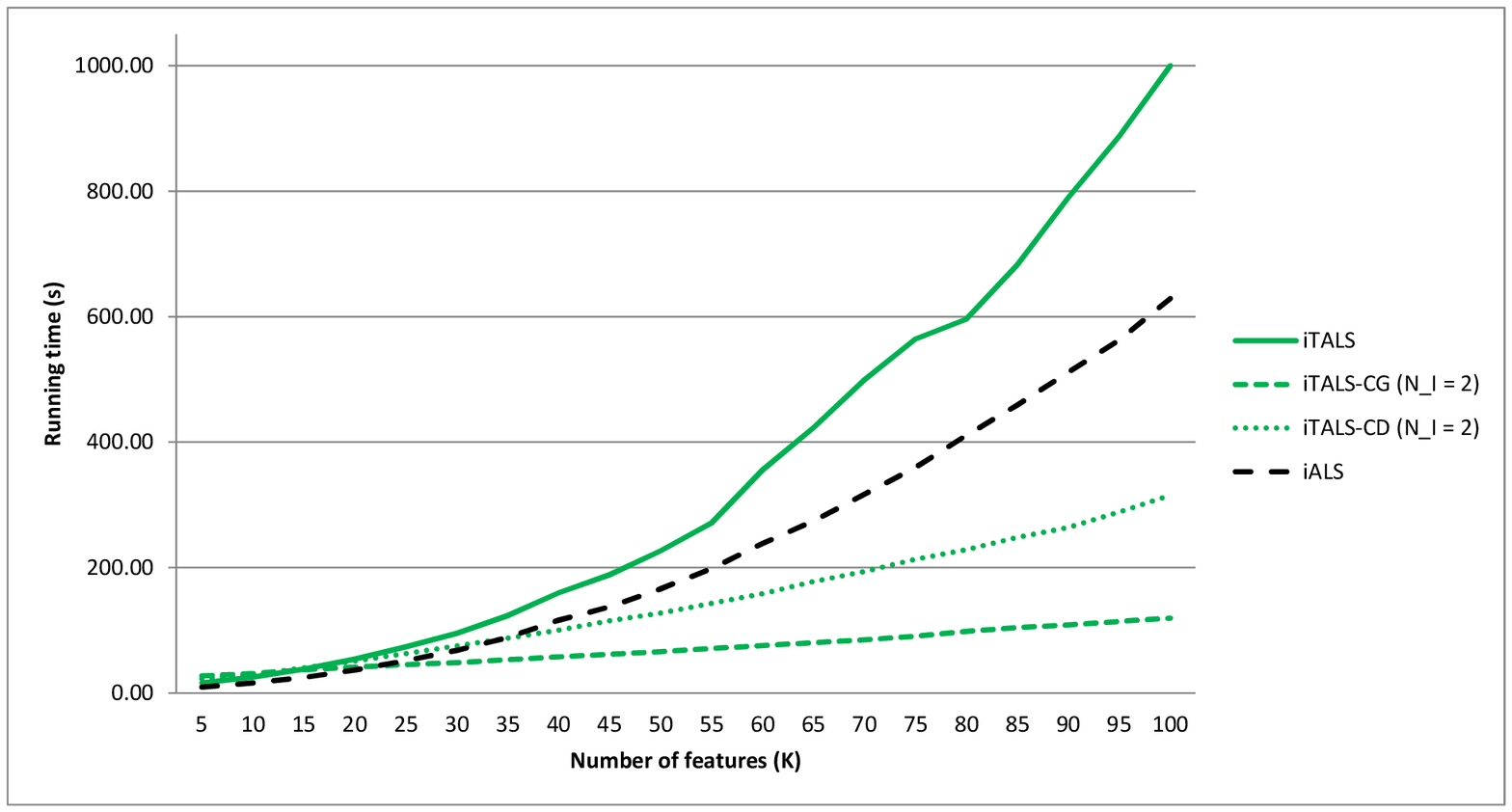}
\caption{Running times of one epoch of iTALS (ALS, CD, CG) and iALS in the value of $K$, using one CPU core}
\label{fig:runtimes}
\end{figure}

Figure~\ref{fig:runtimes} shows running time for each learning method on the LastFM data set for different $K$ values with seasonality as context (using sequentiality results a similar graph). We measured the running time of 10 epochs and use the average of these experiments on the figure. The results are also compared with the running time of iALS. The iALS and iTALS methods appear to scale quadratically with $K$ in practice as we expected (for ``smaller'' $K$ values), CG approximations scale linearly in $K$ and CD is somewhere in between the two. The original iTALS takes 1.5 times more time to learn than iALS, but this means that the time of learning one feature matrix is basically the same for both methods (iTALS learns 3 matrices as opposed to the 2 matrices for iALS). We can also note that both CG and CD increases scalability of the algorithm significantly. The achievable speed-up depends on the number of features ($K$) as well. It is considerable and especially important with larger $K$ values and thus makes the training of high factor models possible in practical applications.

\begin{figure}[!h]
\centering
\includegraphics[width=\hsize]{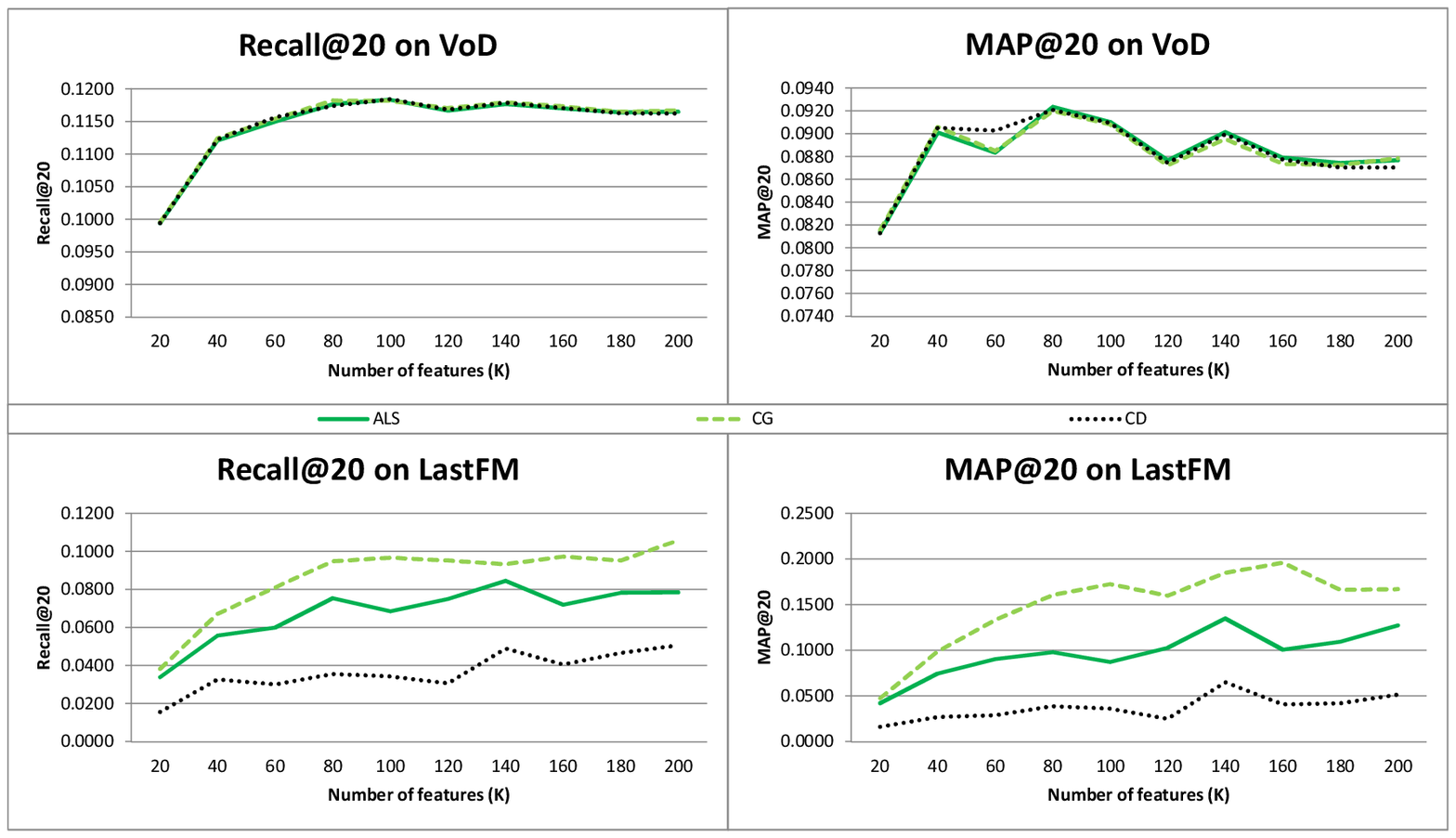}
\caption{Recall@20 (left) and MAP@20 (right) values for iTALS using ALS, CD and CG in the value of $K$.}
\label{fig:factor}
\end{figure}

The approximate solvers can use information from the previous epoch (unlike LS) as the initial $x_0$ solution can be set to the previous value of the feature vectors. This property prevents them from degrading (compared to the ALS based variant) as the number of features increases, even if the number of inner iterations is set to a small fixed value.

Figure~\ref{fig:factor} shows recall@20 (left) and MAP@20 (right) values for ALS, CD and CG based training in the number of features on VoD (up) and LastFM (down) using seasonality. The number of inner iterations for the approximate methods is 2. The graphs for VoD are basically the same for all three training methods. The results of CD on the LastFM dataset are significantly worse than that of the ALS, while CG outperforms both of them.

We found that CD sometimes is unstable with the three way tensor factorization, resulting in diverging (e.g. infinite or NaN) feature values\footnote{Experiments with other models -- like the vanilla MF -- suggest that this problem of the CD is specific to the three way model only.}. This behavior is more common when higher regularization coefficients are used along with higher number of factors. Therefore we had to lower the regularization coefficients for some experiments with CD, including the experiments on LastFM when $K>20$. This however produced significantly worse results. This unpredictable/unstable behavior makes CD learning a less attractive option.

It is uncommon that CG significantly outperforms ALS like it did on LastFM with seasonality as the context. This was the only setting amongst our experiments on which it happened, generally it achieved similar results to ALS. The LastFM dataset has lots of duplicate events (same user listens to the same item) and we only use 6 context-states for seasonality. Therefore this special scenario can be too restrictive and hard to learn. However CG can use the value of the feature vectors from the previous epoch during their recomputation, thus it can become gradually better\footnote{If the previous value of the feature vectors was not carried over between epochs, CG generally underperforms ALS.}.

We found that CG generally produces similar results to ALS and is also the most scalable from the three training methods. It can sometimes also outperform ALS on more restrictive problems. CD is middle road in terms of scalability between CG and ALS. However its instability with the three way model in certain settings basically discards it as a practically useful solution. Therefore we chose the CG learning as the default training method for the iTALS algorithm, with the number of inner iterations set to 2. 
\section{Conclusion and future work}\label{sec:conc}
In this paper we presented an efficient ALS-based tensor factorization method for the context-aware implicit feedback recommendation problem. Our method, coined iTALS, scales linearly with the number of \emph{non-zeroes} in the tensor, thus it works well on implicit data. In addition, we presented two variants implementing faster, approximate learning scheme. We also showed two specific examples for context-aware implicit scenario with iTALS. When using the seasonality as context, we efficiently segmented periodical user behavior in different time bands. When exploiting sequentiality in the data, the model was able to tell apart items having different repetitiveness in usage pattern. These variants of iTALS allow us to analyze user behavior by integrating arbitrary contextual information within the well-known factorization framework. Experiments performed on six data sets show that proposed algorithms can greatly improve the performance. Compared to the context-unaware and context-aware baselines iALS and iCA, our algorithm improved recall@20 and MAP@20 significantly in almost all cases, while it is on par with the implicit version of the context-aware FM.

Comparing the iTALS-variants from the accuracy--training time aspects, we concluded that the CG variant offers fast training time (10--50\% of the ALS) and high accuracy using only a few inner iterations. The number of inner iterations can be fixed to a low value, irrespective of the number of features, as CG can use information from previous epochs and thus can keep up with the accuracy of the ALS. SGD learning is not applicable with our method, however it is used by some other algorithms. iTALS-CG has an advantage over SGD based methods in terms of scalability as it can be easily parallelized.  

Our work opens up a new paths for context-aware recommendations for the implicit feedback problem. Future work will include the characterization of problems when $D$-way (iTALS) or pairwise (FM) models are more advantageous, learning the optimal context states from data, as well as the relation between reweighting, context features and the number of features ($K$).

\section*{Appendix}
\vspace*{-5mm}
\begin{algorithm}[!h]
    \caption{Weighted coordinate descent method}\label{alg:cd}
    \textbf{Input:} {$A$: $N_E\times K$ matrix of input examples; $b$: output for the examples; $x^{(0)}$: initial solution; $w$: vector of weights; $\lambda$: regularization coefficient; $N_I$: number of iterations} \newline
    \textbf{Output:} {$x$: approximate solution of $Ax=b$} \newline
    \textbf{procedure} \Call{Solve-weighted-CD}{$A$, $b$, $x_0$, $w$, $\lambda$, $N_I$}
	\begin{algorithmic}[1]
        \State $x \leftarrow x^{(0)}$
        \For{$i=1,\ldots,N_I$}
            \For{$j=1,\ldots,N_E$}
                \State $\Delta x_j \leftarrow \frac{A_{j}^T(w\circ b)-\lambda x_j}{A_{j}^T(w\circ A_{j})+\lambda}$
                \State $x_j \leftarrow x_j+\delta x_j$
            \EndFor
        \EndFor
        \State \textbf{return} $x$
    \end{algorithmic}
    \textbf{end procedure}
\end{algorithm}

\vspace*{-5mm}
\begin{algorithm}[!h]
    \caption{Conjugate gradient method}\label{alg:cg}
    \textbf{Input:} {$A$: $K\times K$ symmetric positive definite matrix; $b$: output vector; $x^{(0)}$: initial solution; $M$: preconditioning matrix (e.g.: $\mathrm{diag}(A)$); $N_I$: number of iterations} \newline
    \textbf{Output:} {$x$: approximate solution of $Ax=b$} \newline
    \textbf{procedure} \Call{solveCG}{$A$, $b$, $x_0$, $M$, $N_I$}
	\begin{algorithmic}[1]
        \State $r^{(0)} \leftarrow b-Ax^{(0)}$
        \State $z^{(0)} \leftarrow M^{-1}r^{(0)}$
        \State $p^{(0)} \leftarrow z^{(0)}$
        \For{$i=0,\ldots,N_I-1$}
            \State $\alpha^{(i)} \leftarrow \frac{(r^{(i)})^Tz^{(i)}}{(p^{(i)})^TAp^{(i)}}$
            \State $x^{(i+1)} \leftarrow x^{(i)}+\alpha^{(i)}p^{(i)}$
            \State $r^{(i+1)} \leftarrow r^{(i)}-\alpha^{(i)}Ap^{(i)}$
            \State $z^{(i+1)} \leftarrow M^{-1}r^{(i+1)}$
            \State $\beta^{(i)} \leftarrow \frac{(z^{(i+1)})^Tr^{(i+1)}}{(z^{(i)})^Tr^{(i)}}$
            \State $p^{(i+1)} \leftarrow z^{(i+1)}+\beta_ip_i$
        \EndFor
        \State \textbf{return} $x^{(N_I)}$
    \end{algorithmic}
    \textbf{end procedure}
\end{algorithm}
\vspace*{-5mm}

\pagebreak
\small
\bibliographystyle{splncs}
\bibliography{citations}

\begin{thebibliography}{10}

\bibitem{itals_ecml}
Hidasi, B., Tikk, D.:
\newblock Fast {ALS}-based tensor factorization for context-aware
  recommendation from implicit feedback.
\newblock In: Proc. of the ECML-PKDD, Part II. Number 7524 in LNCS.
\newblock Springer (2012)  67–--82

\bibitem{LopsRSH11}
Lops, P., Gemmis, M., Semeraro, G.:
\newblock Content-based recommender systems: State of the art and trends.
\newblock In Ricci, F., Rokach, L., Shapira, B., Kantor, P.B., eds.:
  Recommender Systems Handbook.
\newblock Springer (2011)  73--105

\bibitem{PilaRecsys09}
Pil\'aszy, I., Tikk, D.:
\newblock Recommending new movies: Even a few ratings are more valuable than
  metadata.
\newblock In: Recsys'09: ACM Conf. on Recommender Systems, New York, NY, USA
  (2009)  93--100

\bibitem{Netflix}
Bennett, J., Lanning, S.:
\newblock {The Netflix Prize}.
\newblock In: KDD Cup Workshop at SIGKDD'07, San Jose, California, USA (2007)
  3--6

\bibitem{BellkorICDM07}
Bell, R.M., Koren, Y.:
\newblock {Scalable collaborative filtering with jointly derived neighborhood
  interpolation weights}.
\newblock In: ICDM'07: IEEE Int. Conf. on Data Mining, Omaha, NE, USA (2007)
  43--52

\bibitem{brismf}
Tak\'{a}cs, G., Pil\'{a}szy, I., N\'{e}meth, B., Tikk, D.:
\newblock Major components of the {Gravity} recommendation system.
\newblock SIGKDD Explor. Newsl. \textbf{9} (December 2007)  80--83

\bibitem{Recsys10}
Pil\'aszy, I., Zibriczky, D., Tikk, D.:
\newblock Fast {ALS}-based matrix factorization for explicit and implicit
  feedback datasets.
\newblock In: Recsys'10: ACM Conf. on Recommender Systems, Barcelona, Spain
  (2010)  71--78

\bibitem{Recsys11}
Tak\'{a}cs, G., Pil\'{a}szy, I., Tikk, D.:
\newblock Applications of the conjugate gradient method for implicit feedback
  collaborative filtering.
\newblock In: RecSys'11: ACM Conf. on Recommender Systems, Chicago, IL, USA
  (2011)  297--300

\bibitem{KorenKDD08}
Koren, Y.:
\newblock Factorization meets the neighborhood: a multifaceted collaborative
  filtering model.
\newblock In: SIGKDD'08: ACM Int. Conf. on Knowledge Discovery and Data Mining,
  Las Vegas, Nevada, USA (2008)  426--434

\bibitem{Salak08}
Salakhutdinov, R., Mnih, A.:
\newblock Probabilistic matrix factorization.
\newblock In Platt, J.C., Koller, D., Singer, Y., Roweis, S., eds.: Advances in
  Neural Information Processing Systems 20.
\newblock MIT Press, Cambridge, Massachusetts, USA (2008)

\bibitem{RicciRSH11}
Ricci, F., Rokach, L., Shapira, B.:
\newblock Introduction to recommender systems handbook.
\newblock In: Recommender Systems Handbook.
\newblock Springer (2011)  1--35

\bibitem{AdomaviciusRecsys08}
Adomavicius, G., Tuzhilin, A.:
\newblock Context-aware recommender systems.
\newblock In: Recsys'08: ACM Conf. on Recommender Systems, Lausanne,
  Switzerland (2008)  335--336

\bibitem{KaratzogluRecsys10}
Karatzoglou, A., Amatriain, X., Baltrunas, L., Oliver, N.:
\newblock Multiverse recommendation: {N}-dimensional tensor factorization for
  context-aware collaborative filtering.
\newblock In: Recsys'10: ACM Conf. on Recommender Systems, Barcelona, Spain
  (2010)  79--86

\bibitem{RendleSIGIR11}
Rendle, S., Gantner, Z., Freudenthaler, C., Schmidt-Thieme, L.:
\newblock Fast context-aware recommendations with factorization machines.
\newblock In: SIGIR'11: ACM Int. Conf. on Research and Development in
  Information, Beijing, China (2011)  635--644

\bibitem{AdomaviciusACMTIS05}
Adomavicius, G., Sankaranarayanan, R., Sen, S., Tuzhilin, A.:
\newblock Incorporating contextual information in recommender systems using a
  multidimensional approach.
\newblock ACM Trans. Inf. Syst. \textbf{23}(1) (2005)  103--145

\bibitem{CARS2009}
Adomavicius, G., Ricci, F.:
\newblock Workshop on context-aware recommender systems {(CARS-2009)}.
\newblock In: Recsys'09: ACM Conf. on Recommender Systems, New York, NY, USA
  (2009)  423--424

\bibitem{CAMRA2010}
Said, A., Berkovsky, S., De~Luca, E.W.:
\newblock Putting things in context: Challenge on context-aware movie
  recommendation.
\newblock In: CAMRa'10: Workshop on Context-Aware Movie Recommendation,
  Barcelona, Spain (2010)  2--6

\bibitem{BogersCARS10}
Bogers, T.:
\newblock Movie recommendation using random walks over the contextual graph.
\newblock In: CARS'10: 2nd Workshop on Context-Aware Recommender Systems,
  Barcelona, Spain (2010)  1--5

\bibitem{BaltrunasCARS09}
Baltrunas, L., Amatriain, X.:
\newblock Towards time-dependant recommendation based on implicit feedback.
\newblock In: CARS'09: Workshop on Context-aware Recommender Systems, New York,
  NY, USA (2009)  1--5

\bibitem{BaderCRR11}
Bader, R., Neufeld, E., Woerndl, W., Prinz, V.:
\newblock Context-aware {POI} recommendations in an automotive scenario using
  multi-criteria decision making methods.
\newblock In: CaRR'11: Workshop on Context-awareness in Retrieval and
  Recommendation, Palo Alto, CA, USA (2011)  23--30

\bibitem{HeWWW10}
He, Q., Pei, J., Kifer, D., Mitra, P., Giles, L.:
\newblock Context-aware citation recommendation.
\newblock In: WWW'10: Int. Conf. on World Wide Web, Raleigh, NC, USA (2010)
  421--430

\bibitem{PannielloRecsys09}
Panniello, U., Tuzhilin, A., Gorgoglione, M., Palmisano, C., Pedone, A.:
\newblock Experimental comparison of pre- vs. post-filtering approaches in
  context-aware recommender systems.
\newblock In: Recsys'09: ACM Conf. on Recommender Systems, New York, NY, USA
  (2009)  265--268

\bibitem{hosvd}
Lathauwer, L.D., Moor, B.D., Vandewalle, J.:
\newblock A multilinear singular value decomposition.
\newblock SIAM J. Matrix Anal. Appl. \textbf{21}(4) (2000)  1253--1278

\bibitem{HuICDM08}
Hu, Y., Koren, Y., Volinsky, C.:
\newblock Collaborative filtering for implicit feedback datasets.
\newblock In: ICDM-08: IEEE Int. Conf. on Data Mining, Pisa, Italy (2008)
  263--272

\bibitem{one_class_cf}
Pan, R., Zhou, Y., Cao, B., Liu, N.N., Lukose, R.M., Scholz, M., Yang, Q.:
\newblock One-class collaborative filtering.
\newblock In: ICMD'08: 8$^{\rm th}$ IEEE Int. Conf. on Data Mining, Pisa, Italy
  (2008)  502--511

\bibitem{commendo_track2}
Jahrer, M., T{\"o}scher, A.:
\newblock Collaborative filtering ensemble for ranking.
\newblock In: KDD Cup Workshop at 17$^{th}$ ACM SIGKDD Int. Conf. on Knowledge
  Discovery and Data Mining, San Diego, CA, USA (2011)

\bibitem{Recsys12}
Tak\'acs, G., Tikk, D.:
\newblock Alternating least squares for personalized ranking.
\newblock In: Recsys'12: 6th ACM Conf. on Recommender Systems, Dublin, Ireland
  (2012)  83--90

\bibitem{bpr}
Rendle, S., Freudenthaler, C., Gantner, Z., Schmidt-Thieme, L.:
\newblock Bpr: Bayesian personalized ranking from implicit feedback.
\newblock In: UAI '09: 25$^{th}$ Conf. on Uncertainty in Artificial
  Intelligence. (2009)  452--461

\bibitem{Rendle12}
Rendle, S.:
\newblock Factorization machines with libfm.
\newblock ACM Transactions on Intelligent Systems and Technology (TIST)
  \textbf{3}(3) (2012) ~57

\bibitem{cg}
Hestenes, M., Stiefel, E.:
\newblock Methods of conjugate gradients for solving linear systems.
\newblock Journal of Research of the National Bureau of Standards (1952)
  409--436

\bibitem{LiuCAMRA10}
Liu, N.N., Cao, B., Zhao, M., Yang, Q.:
\newblock Adapting neighborhood and matrix factorization models for context
  aware recommendation.
\newblock In: CAMRa'10: Workshop on Context-Aware Movie Recommendation,
  Barcelona, Spain (2010)  7--13

\bibitem{Agrawal93}
Agrawal, R., Imieli\'{n}ski, T., Swami, A.:
\newblock Mining association rules between sets of items in large databases.
\newblock In: SIGMOD'93: ACM SIGMOD Int. Conf. on Management of Data,
  Washington DC, USA (1993)  207--216

\bibitem{JamesRecsys10}
Davidson, J., Liebald, B., Liu, J.,  et~al.:
\newblock {The YouTube video recommendation system}.
\newblock In: Recsys'10: ACM Conf. on Recommender Systems, Barcelona, Spain
  (2010)  293--296

\bibitem{lastfm1k}
Celma, O.:
\newblock {Music Recommendation and Discovery in the Long Tail}.
\newblock Springer (2010)

\bibitem{tv1_tv2}
Cremonesi, P., Turrin, R.:
\newblock Analysis of cold-start recommendations in iptv systems.
\newblock In: Proceedings of the 2009 ACM Conference on Recommender Systems.
  (2009)

\bibitem{Movielens}
{GroupLens Research}:
\newblock Movielens data sets (2006) \url{http://www.grouplens.org/node/73}.

\end{thebibliography}

\end{document}